\documentclass[10pt,twocolumn,letterpaper]{article}

\usepackage{cvpr}              

\usepackage{graphicx}
\usepackage{amsmath}
\usepackage{amssymb}
\usepackage{booktabs}

\usepackage[table]{xcolor}

\usepackage[pagebackref,breaklinks,colorlinks]{hyperref}

\usepackage[capitalize]{cleveref}
\crefname{section}{Sec.}{Secs.}
\Crefname{section}{Section}{Sections}
\Crefname{table}{Table}{Tables}
\crefname{table}{Tab.}{Tabs.}

\def\cvprPaperID{*****} 
\def\confName{CVPR}
\def\confYear{2022}

\begin{document}

\title{CMR3D: Contextualized Multi-Stage Refinement for 3D Object Detection}

\author{Dhanalaxmi Gaddam$^{1}$
\and
Jean Lahoud$^{1}$
\and
Fahad Shahbaz Khan$^{1,2}$
\and
Rao Muhammad Anwer$^{1,3}$
\and
Hisham Cholakkal$^{1}$
\and
{\tt\small \{Dhanalaxmi.Gaddam,jean.lahoud,fahad.khan,rao.anwer,hisham.cholakkal\}@mbzuai.ac.ae}
\and
$^{1}$ Mohamed Bin Zayed University of Artificial Intelligence, Abu Dhabi, UAE.\\
$^2$ Linköping University, Sweden.\\
$^3$  Aalto University, Finland.
}
\maketitle

\begin{abstract}
   Existing deep learning-based 3D object detectors typically rely on the appearance of individual objects and do not explicitly pay attention to the rich contextual information of the scene. 
  In this work, we propose Contextualized Multi-Stage Refinement for 3D Object Detection (CMR3D) framework, which takes a 3D scene as input and strives to explicitly integrate useful contextual information of the scene at multiple levels to predict a set of object bounding-boxes along with their corresponding semantic labels. 
  To this end, we propose to utilize a context enhancement network that captures the contextual information at different levels of granularity followed by a multi-stage refinement module to progressively refine  the box positions and class predictions.  Extensive experiments on the large-scale ScanNetV2 benchmark reveal the benefits of our proposed method, leading to an absolute improvement of 2.0\% over the baseline. In addition to 3D object detection, we investigate the effectiveness of our CMR3D framework for the problem of 3D object counting. Our source code will be publicly released.
\end{abstract}

\section{Introduction}
3D object detection from point clouds is an emerging computer vision technique that has a diverse set of applications, which include autonomous driving, robotic navigation, and augmented reality, among many others. Object detection methods aim to simultaneously localize and classify 3D objects from a 3D point set. Compared to 2D object detection on 2D images, 3D object detection on point clouds is more challenging due to the increased dimensionality of the object search space and the order-less and sparse nature of point clouds.


Recently, several methods \cite{qi2019deep,xie2020mlcvnet,qi2020imvotenet,xie2021venet} propose to  detect 3D objects directly from point clouds. These methods generally utilize 3D point processing backbone networks \cite{shi2019pointrcnn,qi2017pointnet++} to capture geometric information from raw point clouds followed by a Hough voting mechanism \cite{qi2019deep} to estimate object centers. Features encoded on these object centers are then used to predict the 3D bounding boxes.  Among these voting-based approaches, H3DNet \cite{zhang2020h3dnet} achieves superior performance by  introducing an intermediate representation comprising a hybrid and overcomplete set of geometric primitives. The geometric primitives used in H3DNet are bounding box centers, bounding box 2D face centers, and bounding box edge centers. Although H3DNet achieves promising results, it does not have an explicit mechanism to  incorporate rich contextual information which is helpful for 3D object detection. 

\begin{figure*}
    \centering
    \begin{tabular}{cccc}     
    \textbf{Baseline} & \textbf{Context-H3D} & \textbf{CMR3D} & \textbf{Ground Truth}\\
    \includegraphics[width=0.24\textwidth]{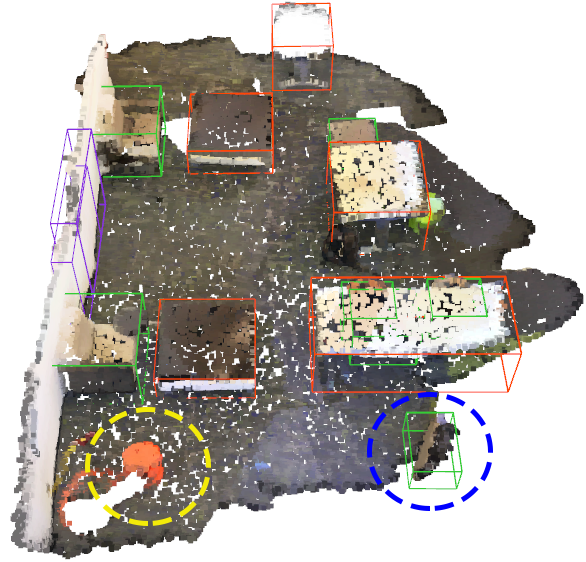}
        &
        \includegraphics[width=0.24\textwidth]{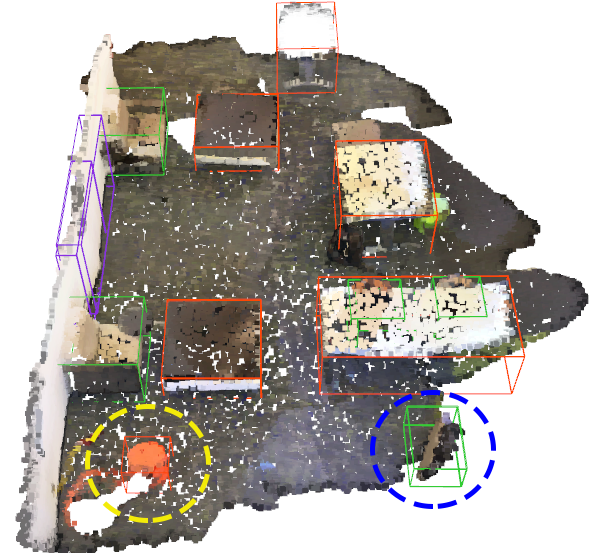}
        &
         \includegraphics[width=0.24\textwidth]{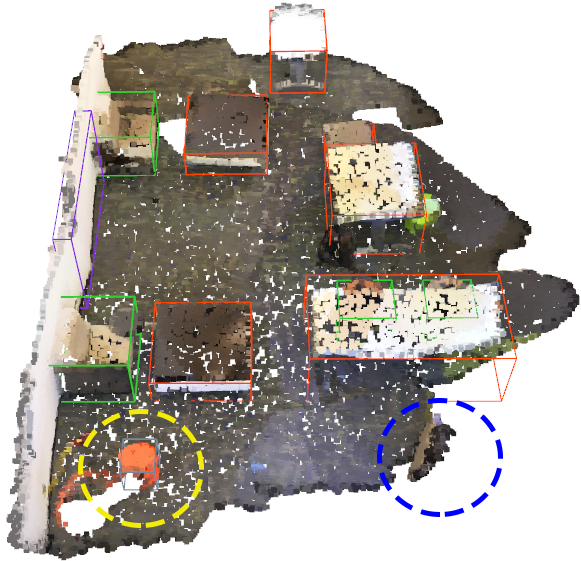}
         &
        \includegraphics[width=0.24\textwidth]{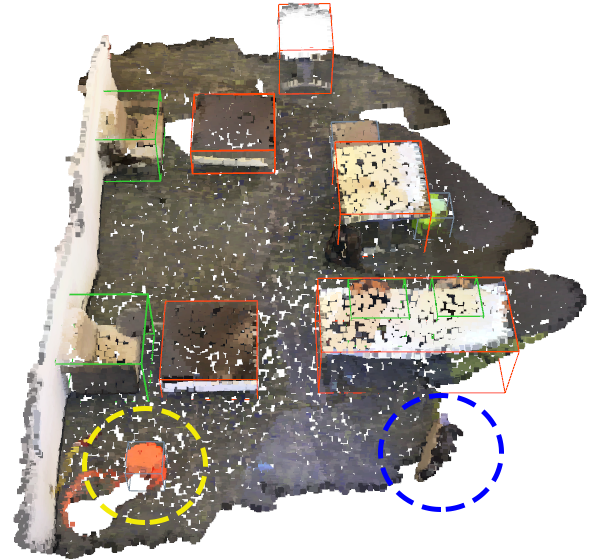}\\
        \includegraphics[width=0.2\textwidth]{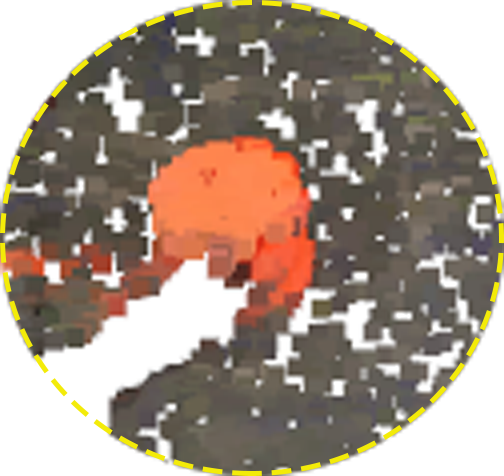}&
        \includegraphics[width=0.2\textwidth]{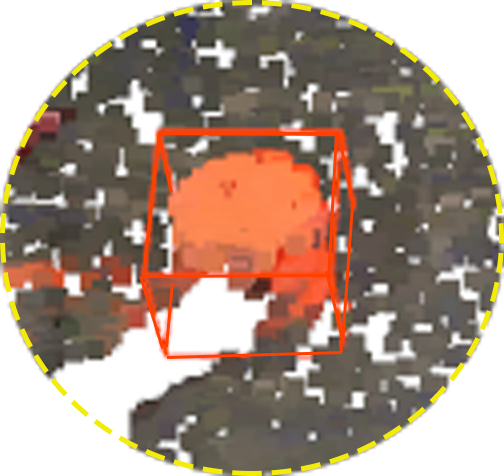}&
        \includegraphics[width=0.2\textwidth]{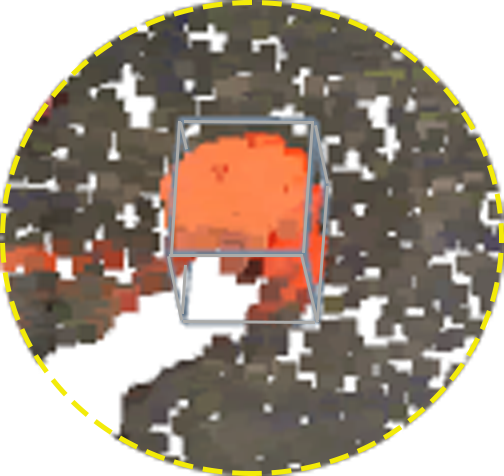}&
        \includegraphics[width=0.2\textwidth]{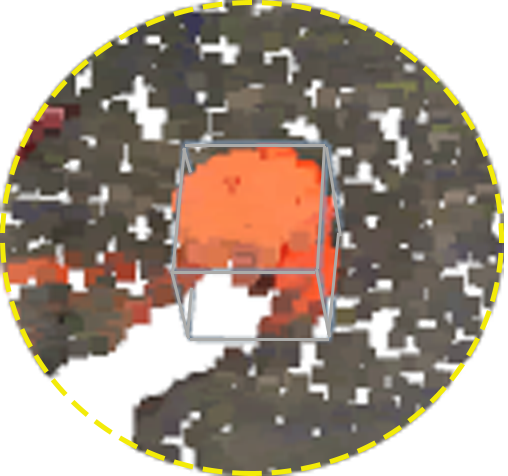}\\
        
    \end{tabular}
    
    \caption{Comparison of 3D object detection results on  ScanNetV2 dataset.    Progressively integrating our context enhancement network (Context-H3D) and multi-stage refinement modules (CMR3D) into the baseline (column 1) leads to improved detection results.   The regions inside the yellow circle are enlarged and shown in the second row for better visualization.  Integration of contextual information helps to localize objects under challenging scenarios (yellow circle). Our multi-stage refinement strategy helps to  refine  the  class predictions (yellow circle) and refine box predictions (blue circle where the false predictions are removed). Note that point clouds are colored only for illustration and not utilized in the input to our proposed method.}
    \label{fig:all_models}
\end{figure*}

In this work, we introduce a context enhancement network that improves the baseline H3DNet by capturing contextual information at different granularity levels. Our context enhancement network consists of three modules: (1) geometric  context module, (2) proposal context module, and (3) hybrid context module, for tightly integrating  contextual information into the geometric primitive and  proposal generation modules of the baseline. Furthermore, motivated by Cascade R-CNN \cite{cai2018cascade} 2D detector, we introduce a multi-stage refinement strategy composed of a sequence of detectors that are trained with increasing threshold values to progressively refine the 3D bounding box locations and class predictions (see Fig. \ref{fig:all_models}).  
The main contributions of our work are the following:
\begin{itemize}
    \item We propose a novel 3D object detection framework named   Contextualized Multi-Stage Refinement 3D Object Detector  (CMR3D) comprising a context enhancement network and  a multi-stage refinement module.  Our CMR3D is constituted of three modules, namely  geometric  context module, proposal  context module, and hybrid context module for  integrating various levels of contextual information into the baseline. 
    Our geometric and  proposal context modules employ self-attention layers, while the hybrid context module employs a multi-scale feature fusion layer.  To  further improve the detection results,  we introduce a  multi-stage refinement module that progressively refines the 3D bounding box locations and class predictions.
    
         \item Extensive experiments on the challenging ScanNetV2 dataset demonstrate the benefits of our proposed method. While using the same PointNet++ backbone, CMR3D performs favorably against the baseline with an  absolute gain of 2\%.
         \item We extend our 3D object detection framework, CMR3D,  for multi-category object counting on point clouds. To the best of our knowledge, we are the first to investigate the problem of multi-category object counting on point clouds. Our object counting results on ScanNetV2 show that the proposed context enhancement network and multi-stage refinement module lead to improved counting results over the baseline. \end{itemize}
\begin{figure*}
    \centering
    \includegraphics[width=\textwidth]{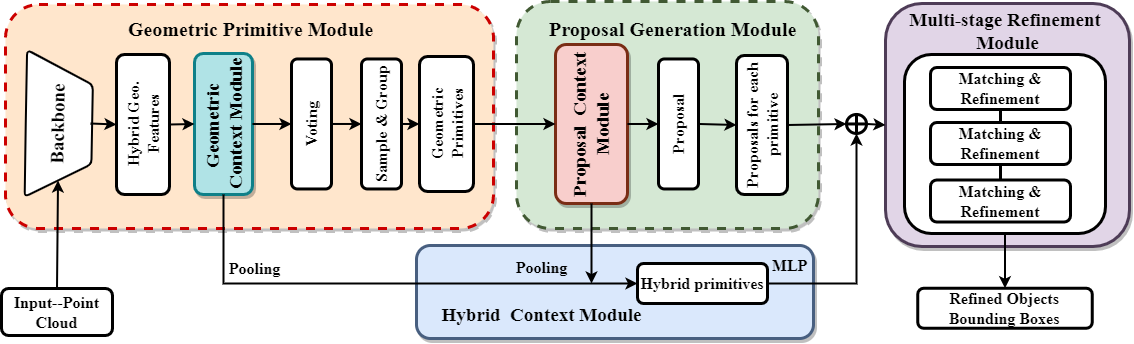}
    \caption{Overview of the proposed contextualized 3D object detector CMR3D.  The geometric context module (GCM) and proposal context module (PCM) are introduced in the geometric primitive module and proposal generation module respectively. GCM takes in primitives and outputs the primitives with the relationship between its neighboring primitives. PCM takes these geometric primitives and provides the object proposals for each primitive. The hybrid context module (HCM) utilizes the primitives of GCM and PCM for understanding the whole scene. The multi-stage refinement module helps in refining the proposals obtained and outputs the 3D bounding boxes.}
    \label{fig:architecture}
\end{figure*}
\section{Related Work }
We propose a 3D object detection model that integrates the contextual information at different granularity levels and utilizes the multi-stage refinement module which refines the predicted bounding boxes and class predictions.
\subsection{3D Object Detection}
Given the advancement of deep learning methods on images and the abundance of 2D datasets, some 3D object detection methods choose to learn scene representation by projecting into a 2D plane.  MV3D \cite{engelmann20203d} and VoxelNet \cite{zhou2018voxelnet} use 3D data which is reduced to a 2D bird's eye view and then fed to the network. These approaches depend on 2D detectors, which restricts the model from learning 3D shape information from point clouds. As a solution to this problem, many methods propose to directly process point clouds. 3D-SIS \cite{hou20193d} and 3D-BoNet \cite{yang2019learning} use 3D instance segmentation methods to learn object bounding boxes. PointRCNN \cite{shi2019pointrcnn} presents a two-stage 3D object detector that generates the bounding box proposals first and then refines them to achieve the detection results.
Motivated by the Hough Voting mechanism \cite{leibe2004combined} in 2D object detection, VoteNet \cite{qi2019deep} proposes learning to vote towards object centers. It uses PointNet++ \cite{qi2017pointnet++} as a backbone for feature extraction and vote estimation, and it then detects bounding boxes by feature sampling and grouping.

VoteNet is used as a building block for many methods. 3D-MPA \cite{engelmann20203d} utilizes the voting mechanism and graph convolution networks for refining object proposals and uses 3D geometric features for aggregating object detections. MLCVNet \cite{xie2020mlcvnet} improves VoteNet by using multi-level contextual information for object candidates. HGNet \cite{chen2020hierarchical} improves the VoteNet and uses a hierarchical graph network with feature pyramids. H3DNet \cite{zhang2020h3dnet} incorporates the votes to additional 3D primitives such as centers of box edges and surfaces. BRNet \cite{cheng2021back} uses the back-tracing operation in Hough voting by querying neighboring points around the object candidates. Modeling the relationship between object proposals by graph structure is demonstrated in RGNet \cite{feng2020relation}. 
Given the challenging real world scenarios, we believe that knowledge of the complete scene would help in further improving the detection accuracy. This is achieved by understanding the relationship between the objects and their surroundings for better scene understanding.

\subsection{Contextual Information}
Contextual information is widely used in improving the performance of many tasks, which include 2D object detection \cite{yu2016role,hu2018relation,liu2018structure}, 3D point matching \cite{deng2018ppfnet}, point cloud semantic segmentation \cite{ye20183d}, and 3D scene understanding \cite{zhang2017deepcontext}. The point patch context is used in 3D point clouds for instance segmentation  \cite{hu2018semantic} and achieves enhanced performance. Hierarchical context priors are introduced in a recursive auto-encoder network \cite{shi2019hierarchy} for detecting 3D objects. Self-attention is widely used in natural language processing \cite{vaswani2017attention} and learns global context. Inspired by this mechanism, contextual information is being used in many applications, such as image recognition \cite{hu2018squeeze}, semantic segmentation \cite{fu2019dual}, and point cloud recognition \cite{xie2018attentional}. PCAN \cite{zhang2019pcan} proposes to utilize the attention network in 3D point data processing for capturing contextual information. Instead of searching the whole input point cloud data, Attentional-PointNet \cite{paigwar2019attentional} suggests searching in regions of interest while detecting 3D objects.
MLCVNet \cite{xie2020mlcvnet} adopts a multi-level contextual information module for 3D object detection from point clouds. In this work, we utilize the concept of multi-level contextual information by integrating two self-attention modules and one multi-scale feature fusion module into a baseline method to learn multi-level contextual relationships between the geometric primitives, object centers, and the complete scene with their surroundings. Also, we use the multi-stage refinement strategy for refining the detected 3D bounding boxes and class predictions.

\textbf{Counting:}
Counting objects is an interesting problem in the research field of computer vision as it helps in managing many scenarios such as traffic control, crowd management, and many more. Several methods \cite{zhang2015cross,segui2015learning} propose to count objects from 2D images; however, to the best of our knowledge, there are few in the 3D domain, especially on 3D point cloud datasets. Counting objects by detection is investigated in \cite{chattopadhyay2017counting},  where per-class object counting is done. Motivated by \cite{cholakkal2019object}, we perform 3D object counting by detection.

\section{Approach}


The proposed 3D object detector, CMR3D, is shown in Fig. \ref{fig:architecture}. It consists of three contextual modules that are integrated within our baseline \cite{zhang2020h3dnet}, along with a multi-stage refinement module. The three contextual modules are the Geometric  Context Module, the Proposal Context Module, and the Hybrid  Context Module. We describe in detail these contextual modules in the following sections. 
\begin{figure*}
    \centering
    \includegraphics[width=\linewidth]{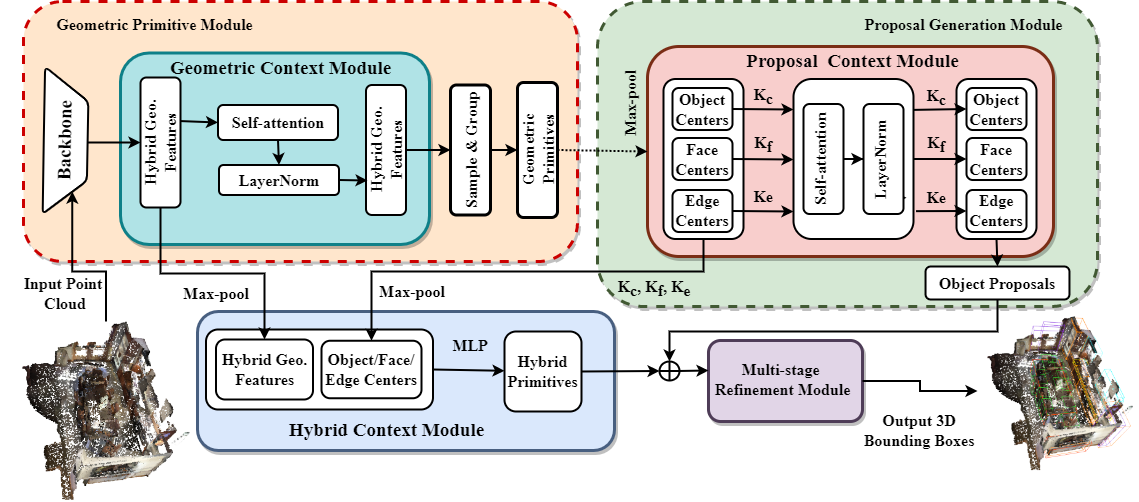}
    \caption{Detailed architecture of the proposed 3D object detector, CMR3D. We introduce a context enhancement network to capture contextual information at three granularity levels  and a multi-stage refinement module which progressively refines the box positions and class prediction. Our context enhancement network tightly integrates contextual information to the baseline through our geometric,  proposal, and hybrid context modules.
   Geometric and proposal context modules use the self-attention layers followed by layer normalization, and the hybrid context module concatenates the proposals using multi-scale feature fusion layer.
    \textit{ Point clouds are colored for  illustration only.}}
    \label{fig:context_modules}
\end{figure*}
\subsection{Baseline Method} \label{h3dnet}

We choose H3DNet \cite{zhang2020h3dnet} as our baseline method. It consists of three main modules: \textit{geometric primitive module, proposal generation module}, and \textit{classification and refinement module}. It uses a PointNet++ backbone, which is a common approach adopted by numerous point-based methods, to learn features for a sub-sampled set of points. Features of downsampled points are used as input to a geometric primitive module that predicts the locations of the bounding boxes (BB) centers, faces, and edge centers. The centers are predicted using a Multi-Layer Perceptron (MLP), adapted from VoteNet's Hough voting. These centers are passed through the proposal generation module, where a distance function is used for optimizing these centers and obtaining local minima which provide high-fidelity object proposals. The final module takes these object proposals as input and classifies them as objects or not. This is achieved by combining the geometric features in the neighborhood of the object proposals with the help of MLP layers.

\textbf{Limitations:} In the baseline method, the geometric primitives are converted into object proposals by defining a distance function between an object and the geometric primitives \cite{zhang2020h3dnet}. 
However, we argue that the complete relationship between these primitives is not achieved and can be observed in Fig. \ref{fig:all_models}. To this end, we propose CMR3D to obtain more information regarding the centers, proposals, and scenes at each level respectively.

\subsection{CMR3D} \label{cmr3d}
To this end, we propose CMR3D, a 3D object detection network that utilizes the contextual information of the scene to better identify the 3D object bounding boxes. This is achieved with the help of three contextual models which are integrated into the baseline method: geometric context module (GCM), proposal context module (PCM), and hybrid context module (HCM). GCM helps the initial primitives (which are taken as input from the backbone) in capturing the information related to its neighboring primitives. This helps in increasing the voting accuracy for obtaining the accurate BB centers, BB face centers, and BB edge centers. PCM is used to get the relationship between the object proposals which are generated from each of the obtained BB centers. Further, HCM helps the network in understanding the relationship between the objects and their surroundings. Finally, the multi-stage refinement module progressively refines the box positions and class predictions. Next, we provide a detailed discussion about our novel modules.  
\subsubsection{Geometric Context Module (GCM)}
It is the first level of the context module introduced for retrieving the missing information within the initial primitives by collecting the necessary data from similar primitives, i.e., BB centers, BB face centers, and BB edge centers. 
Therefore, our GCM sub-module aims to better understand the relationships between geometric primitives to improve the voting accuracy.
The process is repeated with each primitive to enhance the voting accuracy by utilizing the concept of self-attention which can encode long-range dependencies.
The squeeze and excitation method \cite{hu2018squeeze} shows that correlation between the channels promotes in retrieving the contextual information, especially in the object detection tasks. To this end, we use the concept of compact generalized non-local network (CGNL) \cite{yue2018compact} as the self-attention mechanism for exploiting the correlations between primitives. After the extraction of points from PointNet++, we get the feature map $F$.
The feature map $F \in \mathbb{R}^{n \times D}$ represents the hybrid geometric features of $n$ points, which are subsampled from the input point cloud, and $D$ is the feature vector dimension.
The new feature map $F'$ encodes the relationship between the geometric primitives, which can be formulated as:
\begin{equation}
    F' = LN(f(\theta(F), \phi(F))g(F))
    \label{gpcm}
\end{equation}
Here, LN is the LayerNorm operation \cite{ba2016layer} which is applied along the channels of the feature map. 
$\theta(F), \phi(F)$, and $g(F)$ are three different transform functions and $f(\theta(F), \phi(F))$ encodes the similarities between the two primitive positions. 
The new feature map $F'$ is obtained after applying self-attention, and each primitive holds its own primitive features along with the information of other primitives. This whole process is repeated for all three centers i.e., BB center, BB face center, and BB edge center. The aqua-colored region inside the geometric primitive module in Fig. \ref{fig:context_modules} shows the GCM module.

\subsubsection{Proposal Context Module (PCM)}
It is the second-level context module introduced for enhancing  the relationship between the object proposals. 
In the baseline method, the geometric primitives, BB centers, BB face centers, and BB edge centers are directly taken as input for obtaining the object proposals. 
We believe that this does not fully learn the relationships between objects. 
Instead, we propose to first pass the primitives obtained after voting into a self-attention module. By grouping the voted centers for each primitive, we attain a set of clusters for each geometric primitive: $K_c, K_f$, and $K_e$ for BB center, BB face center, and BB edge center respectively. These clusters can be represented as $K = \{k_1,k_2,\hdots,k_N\}$ where $N$ is the number of clusters generated by grouping the predicted centers. We provide the formulation for one $K$, and the same applies for all three $K_c, K_f, K_e$. To make  $k = \{c_1,c_2,\hdots,c_n\}$  as a single vector form representation of cluster, it is fed into an MLP followed by the max-pooling operation. Here, $c_i$ represents the $i-$th vote and $n$ is the number of votes present in each $k$. This is shown in the pink region of architecture, Fig. \ref{fig:context_modules}.

To consider the relationship between these cluster vectors, we pass the cluster features through a self-attention module.
The cluster features $K \in \mathbb{R}^{N \times D'}$ are fed to the attention module to generate a new feature map $K_{P} \in \mathbb{R}^{N \times D'}$ which stores the relation between them, $D'$ is the dimension of the feature map. This can be formulated as:
\begin{equation}
    k_p =  \text{max-pool} \{LN(SA(  MLP(c_i)))\}
\end{equation}
where $k_p$ is the new feature vector of new feature map $K_P$, $LN$ is the layer norm operation and $SA()$ is a self-attention similar to equation \ref{gpcm}.

\subsubsection{Hybrid Context Module (HCM)}
It is the third-level of context module which is introduced to capture the whole scene information to improve the detection accuracy. We believe that having access to information from the entire scene helps the network to easily detect an object and reduce ambiguity. Motivated by the structure inference net \cite{liu2018structure}, we propose the HCM module which enables the network to learn about contextual information and further improve the feature representation for BB proposals.

In order to achieve this, we create a new branch, where geometric primitives and clusters are taken (as shown in the blue region of Fig. \ref{fig:context_modules}) from GCM and PCM modules before they are processed through the self-attention layer. 
Primitives ($\mathcal{G}$) and cluster features ($k$) are then pooled using the max-pool operation, and the resulting feature vectors are concatenated.
Using an MLP, the obtained features are combined, and the output is expanded to match the size of the feature map of PCM module. This increases the BB detection performance as it considers the whole scene information. This operation can be formulated as:
\begin{equation}
    K_h = MLP([\text{max-pool}(k); \text{max-pool} (\mathcal{G})]) + K_P
\end{equation}
The geometric primitives  are $\mathcal{G} \in \textbf{G} = \{\mathcal{G}_1, \mathcal{G}_2, \hdots, \mathcal{G}_P\}$ and object proposals (or clusters)  are $k \in K = \{k_1,k_2,\hdots,k_N\}$ where $P$ and $N$ are sampled number of primitives and clusters respectively. $K_P$ is the outcome of PCM,  $K_h$ is calculated for all three centers i.e, BB centers, BB face centers, and BB edge centers, and are used as input to our last module i.e., multi-stage classification and refinement.

\subsubsection{Multi-stage Refinement Module}
The final module of our network is the multi-stage refinement module which takes object proposals (with rich contextual information) as input and outputs 3D object bounding boxes. This module has both classification and refinement sub-modules. The classification sub-module helps in determining whether the object proposals obtained represent an object or not, and the refinement sub-module determines the offset of detected objects. All the features associated with primitives are combined for each object proposal, as these contain the information which helps in determining the 3D bounding boxes. The baseline method has a single stage for refinement and classification of the predicted BB, whereas we believe that having the multi-stage refinement module enhances the efficiency of the network. Inspired by 2D cascade detection \cite{cai2018cascade}, where different threshold values are used to detect objects at multiple levels. We intend to apply the same mechanism in 3D, where we use three levels of refinement with different threshold values $U=\{0.5, 0.55, 0.6\}$ at each level.

\subsubsection{Extending CMR3D to Object Counting} \label{count}
To the best of our knowledge, this is the first work that proposes counting objects using 3D object detection network, especially on ScanNetV2 \cite{dai2017scannet} dataset. We perform counting of objects by detection, and our main idea is to show that contextual information is helping the detector to better identify object occurrences. 
For counting the objects by detection, we compare our baseline method (H3DNet), Context-H3D, and CMR3D. In order to get the proper count of 3D bounding boxes, we use the objectness score as the confidence level for each category. We choose a threshold confidence score of $0.95$ for all three detectors. After getting the count of objects for each scan in the validation set of ScanNetV2, we use the evaluation criteria similar to the one used in \cite{cholakkal2019object}.
\begin{table}
    \centering
    \resizebox{\linewidth}{!}{
    \begin{tabular}{|l|l|l|c|c|}
    \hline
         \textbf{Method} &  \textbf{Input} & \textbf{BackBone} & \textbf{mAP$@$0.25} & \textbf{mAP$@$0.50}\\
         \hline
         VoteNet & Geo only& PointNet++& 58.6 & 33.5 \\
         MLCVNet & Geo only& PointNet++ & 64.5 & 41.4\\
   \rowcolor{blue!6}    H3DNet & Geo only & PointNet++&64.4 &43.4 \\
           \textbf{CMR3D(Ours)} &\textbf{ Geo only} & \textbf{PointNet++} & \textbf{67.3 (66.5)}& \textbf{48.6 (47.6)}\\
         \hline
  \rowcolor{blue!6}      H3DNet & Geo only & 4$\times$PointNet++&67.2 &48.1 \\
   Context-H3D & Geo only & 4$\times$PointNet++& 67.7 (67.1)& 49.8 (49.4)\\ 
    \textbf{CMR3D(Ours)} & \textbf{Geo only} & \textbf{4$\times$PointNet++}& \textbf{68.1 (67.4)}& \textbf{50.1 (49.7)}\\
         \hline
\end{tabular}}
    \caption{Comparison of 3D object detection results with state-of-the-art methods on ScanNetV2 validation set. CMR3D with single PointNet++ backbone achieves better accuracy compared to H3DNet with four PointNet++ backbones. We show the maximum and (mean) of five runs, and baseline results are highlighted.}
    \label{tab:results}
\end{table}

\begin{table*}
    \centering
    \resizebox{\textwidth}{!}{%
    \begin{tabular}{|l|cccccccccccccccccc|c|}
    \hline
      Method & cab & bed & chair & sofa & tabl & door & windw & bkshf & pic. & contr & desk & curtn & frdge & showr & toilt & sink & bath & ofurn & mAP  \\
      \hline
        
         VoteNet &8.1 &76.1 &67.2 &68.8 &42.4 &15.3 &6.4 &28.0 &1.3 &9.5 &37.5 &11.6 &27.8 &10.0 &86.5 &16.8 &78.9 &11.7 &33.5 \\ 
      MLCVNet &16.6 & 83.3& 78.1& 74.7& 55.1& 28.1& 17.0& 51.7 &3.7 &13.9 &47.7 &28.6 &36.3 &13.4 &70.9 &25.6 &85.7  &27.5 &42.1 \\
\rowcolor{blue!6}        H3DNet &20.5&  79.7& 80.1 &  79.6& 56.2& 29.0& 21.3&45.5 &4.2 &33.5 &50.6 & 37.3& 41.4& 37.0& 89.1& 35.1& 90.2 &35.4 & 48.1\\ 
      CMR3D(\textbf{ours}) & \textbf{22.30}&\textbf{84.46} &\textbf{82.75} &\textbf{81.83} &55.31 &\textbf{38.50} &\textbf{23.13} &43.30 &\textbf{8.83} &29.06 &\textbf{53.43} &34.25 &\textbf{52.25} &\textbf{44.28} &\textbf{94.18} &32.68 &86.57 &35.06 &\textbf{50.1} \\
    \hline
    \end{tabular}}
    \caption{We show category-wise results of mean average precision (mAP) with 3D IoU threshold $0.5$ as proposed in \cite{song2015sun}, and mean of AP across all semantic classes with 3D IoU threshold $0.5$ for 3D object detection on ScanNetV2 validation dataset. The baseline results are highlighted.}
    \label{tab:results_map50}
\end{table*}
 \begin{table*}
     \centering
     \resizebox{\textwidth}{!}{%
    \begin{tabular}{|l|cccccccccccccccccc|c|}
    \hline
      Method & cab & bed & chair & sofa & tabl & door & windw & bkshf & pic. & contr & desk & curtn & frdge & showr & toilt & sink & bath & ofurn & mAP  \\
      \hline
        VoteNet & 36.3 &87.9 & 88.7& 89.6& 58.8& 47.3& 38.1 & 44.6& 7.8& 56.1& 71.7& 47.2& 45.4&57.1 &94.9 & 54.7& 92.1& 37.2 & 58.7\\ 
        MLCVNet & 42.45 & 88.48&  89.98& 87.4&63.50 &56.93 &46.98 &56.94 &11.94 &63.94 &76.05 &56.72 &60.86 &65.91 &98.33 &59.18 &87.22 &47.89 &64.48 \\ 
   \rowcolor{blue!6}  H3DNet & 49.4&88.6 &91.8 &90.2 &64.9 &61.0 &51.9 &54.9 &18.6 &62.0 &75.9 &57.3 &57.2 &75.3 &97.9 &67.4 &92.5 &53.6 &67.2 \\ 
      CMR3D(\textbf{ours}) & \textbf{50.93} &\textbf{90.17} &\textbf{92.09} &88.91 &\textbf{66.12} &60.41 &51.84 &54.20 &18.66 &\textbf{72.54} &\textbf{80.93} &56.63 &51.72 &\textbf{80.56} &96.61 &\textbf{68.52} &89.51 &\textbf{54.52} &\textbf{68.1} \\ 
    \hline
    \end{tabular}}
     \caption{We show category-wise results of mean average precision (mAP) with 3D IoU threshold $0.25$ as proposed in \cite{song2015sun}, and mean of AP across all semantic classes with 3D IoU threshold $0.25$ for 3D object detection on ScanNetV2 validation dataset. The baseline results are highlighted.}
     \label{tab:results_map25}
 \end{table*}
\section{Implementation Details}
Our 3D object detector is implemented in PyTorch \cite{NEURIPS2019_9015}, and it is trained for 360 epochs using an Adam optimizer \cite{kingma2014adam} with batch size 6. The base learning rate is initialized with $1e-2$ and decay steps to $\{80, 140, 200, 240\}$ with decay rates $\{0.1, 0.1, 0.1, 0.1\}$. We trained the model with one Quadro RTX $6000$ GPU, it takes around 22 hours for training. We noticed that while training, the mAP oscillates with a small frequency; therefore, we report the max and mean results of five runs.

\textbf{Dataset}
We use the ScanNetV2 \cite{dai2017scannet} 3D dataset, which contains $1,513$ scans of indoor scenes and annotations for $40$ object categories, of which we consider $18$ categories.
Similar to VoteNet \cite{qi2019deep}, we employ $1,201$ scans for training and $312$ validation scans as test set and sub-sample $40,000$ points from every scan.

\subsection{Evaluation Protocol} \label{evaluation}
Similar to the other 3D object detectors, we use average precision (AP) and the mean average precision (mAP) among all semantic classes \cite{song2015sun} for various IoU (Intersection over union) values. IoU is calculated as the ratio of the area of the intersection to the area of the union of the predicted bounding box and the ground truth bounding box. AP represents the average precision value across recall values ranging from 0 to 1.

  Evaluation for counting is done by using the root mean square error (RMSE) along with its few variants: non-zero RMSE (nz-RMSE), relative RMSE (rRMSE), and non-zero relative RMSE  (nz-rRMSE) \cite{cholakkal2019object}. For a given ground-truth counts $g_{ic}$, predicted counts $p_{ic}$ for a category $c$, and scan $i$, the RMSE is calculated as,
\begin{equation}
    RMSE_c = \sqrt{\frac{1}{N} \sum_{i=1}^N ( p_{ic} - g_{ic})^2}
\end{equation} and relative RMSE is calculated as
\begin{equation}
    rRMSE_c = \sqrt{\frac{1}{N} \sum_{i=1}^N \frac{( p_{ic} - g_{ic})^2}{g_{ic} + 1}}
\end{equation} where $N$ is number of scans present in the validation set. The $m-RMSE$ and $m-rRMSE$ are obtained by calculating an average error across all categories. Similarly, we obtain the count for non-zero ground-truth instances known as $m-nz-RMSE$ and $m-nz-rRMSE$.

\subsection{Results Analysis}
Table \ref{tab:results} shows the results of our 3D object detector, CMR3D, along with other state-of-the-art methods. We partition the table with respect to the number of backbones used and highlight the rows of the baseline method results. CMR3D shows an increase of $0.85\%$ in mAP $0.25$ and $2.02\%$ in mAP $0.50$ when compared to the baseline H3DNet method. 
 Per-category results for 3D IoU threshold of $0.25$ are shown in the supplementary material. 
 We observe that few categories such as picture, fridge, and shower curtain categories have a significant improvement with $4.43, 11.2$, and $7.28$ in mAP $0.50$.
 We also note that from Table \ref{tab:results_map50}, objects that usually co-occur, such as chair, sofa, and cabinet, shower curtain and toilet, window and door show an increase in performance. This shows the capability of the network to better understand the relationship between objects and their surroundings.
 Table \ref{tab:results_map25} shows the class-wise results of our model with the baseline method and previous state-of-the-art methods. It can be noted that, for most of the categories, our model (CMR3D) shows better accuracy compared to the baseline. We can also observe that the objects which are usually seen together such as cabinet, bed, chair (seen mostly together), similarly counter and desk shows better accuracy than the objects that do not appear together. This indicates that the contextual information is helping the model in identifying the closely-related objects more accurately.

 \begin{table}
     \centering
     \resizebox{\linewidth}{!}{
\begin{tabular}{|l|c|c|c|c|}
    \hline
        \textbf{Method} & \textbf{m-RMSE} & \textbf{m-nz-RMSE} &\textbf{ m-rRMSE} & \textbf{m-nz-rRMSE } \\
        \hline
   \rowcolor{blue!6}      H3DNet& 1.1432 & 1.8661 & 0.6102& 0.7971\\
         Context-H3D &1.0803 & 1.7690 & 0.5743 & 0.7635 \\
        \textbf{CMR3D} & \textbf{0.9647} & \textbf{1.6035} & \textbf{0.5097} & \textbf{0.6850} \\
         \hline
    \end{tabular}}
     \caption{Performance comparison for counting objects using the m-RMSE error values with the baseline method, our method without multistage refinement, and our proposed method. Lower values mean better performance. The baseline results are highlighted.}
     \label{tab:rmse}
 \end{table}
 We evaluate the performance of object counting by detection for each category individually. Table \ref{tab:rmse} shows the quantitative results using the mean RMSE values along with it variants. It can be observed that our model performs better compared to the baseline method. We believe that the multi-stage refinement is helping the model to better identify and localize objects. It can be observed that our Context-H3D also performs better than the baseline, which shows the importance of contextual information in enhancing the model's performance. The category-specific RMSE values for the baseline and our method are shown in the supplementary material. Our sole intention to perform the counting on objects is to show that our model can better identify objects compared to the baseline method, and it can also be seen that contextual information is enhancing the model performance. Also, the introduction of multi-stage refinement module helps in improving the accuracy, as can be observed with the low RMSE values. 
\begin{table}
    \centering
    \resizebox{\linewidth}{!}{
\begin{tabular}{|l|c|c|c|c|}
    \hline
        \textbf{GCM} & \textbf{PCM} & \textbf{HCM} &\textbf{ mAP@0.25} & \textbf{mAP@0.50 } \\
        \hline
    \rowcolor{blue!6}    &&& 67.2 & 48.1\\
        \checkmark &  &  &  67.5 & 49.1 \\
         & \checkmark &  &  67.3 & 49.4 \\
        \checkmark & \checkmark  &  & 67.4 & 49.5 \\
         \textbf{\checkmark} & \textbf{\checkmark} & \textbf{\checkmark}   & \textbf{67.7} & \textbf{49.8 }\\
         \hline
    \end{tabular}}
    \caption{Understanding the effectiveness of contextual modules. We have provided the results with individual contextual module also by combining the contextual modules. The baseline results are shown where there are no contextual modules and highlighted baseline results.}
    \label{tab:modules}
\end{table}
\begin{table}
    \centering
    \resizebox{\linewidth}{!}{   
\begin{tabular}{|l|l|l|c|c|}
    \hline
         \textbf{Method} &  \textbf{Input} & \textbf{BackBone} & \textbf{mAP$@$0.25} & \textbf{mAP$@$0.50}\\
         \hline
         
  \rowcolor{blue!6}        H3DNet & Geo only & PointNet++&64.4 &43.4 \\
         Context-H3D & Geo only & PointNet++& 66.4 (65.9)& 48.2 (47.5)\\
         \textbf{CMR3D} & \textbf{Geo only} & \textbf{PointNet++}& \textbf{67.3 (66.5)}& \textbf{48.6 (47.6)}\\
         \hline
    \end{tabular}}
    \caption{Ablation study results on the validation set of ScanNetV2 dataset. We report max and (mean) of the five runs. Single backbone network shows a significant improvement in both our models compared with baseline. The baseline results are highlighted.}
    \label{tab:ablation}
\end{table}
\begin{figure*}
    \centering
    \begin{minipage}{.7\textwidth}
    \centering
    \begin{tabular}{cccc}     
    \textbf{Baseline} & \textbf{Context-H3D} & \textbf{CMR3D} & \textbf{Ground Truth}\\
    \includegraphics[width=0.2\linewidth]{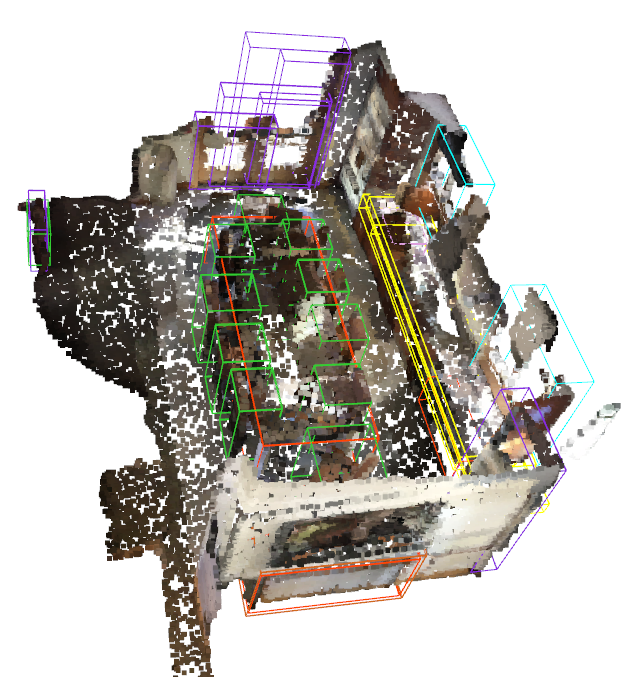}
        &
        \includegraphics[width=0.2\linewidth]{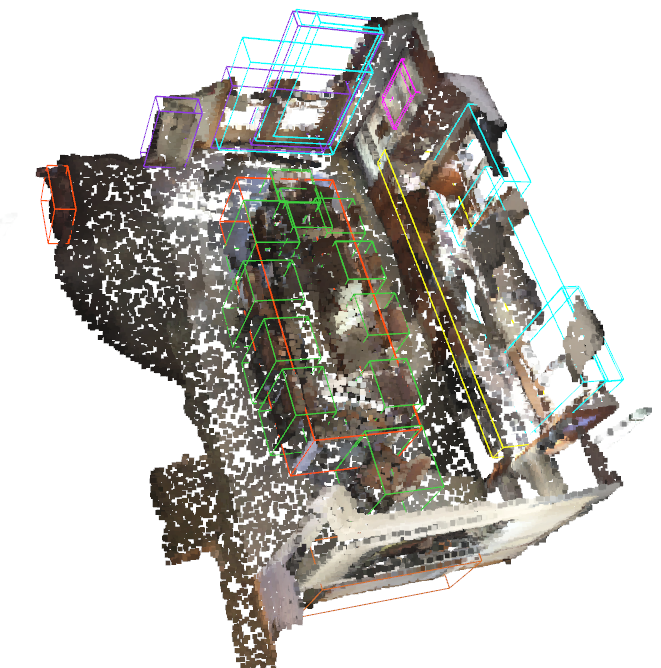}
        &
         \includegraphics[width=0.2\linewidth]{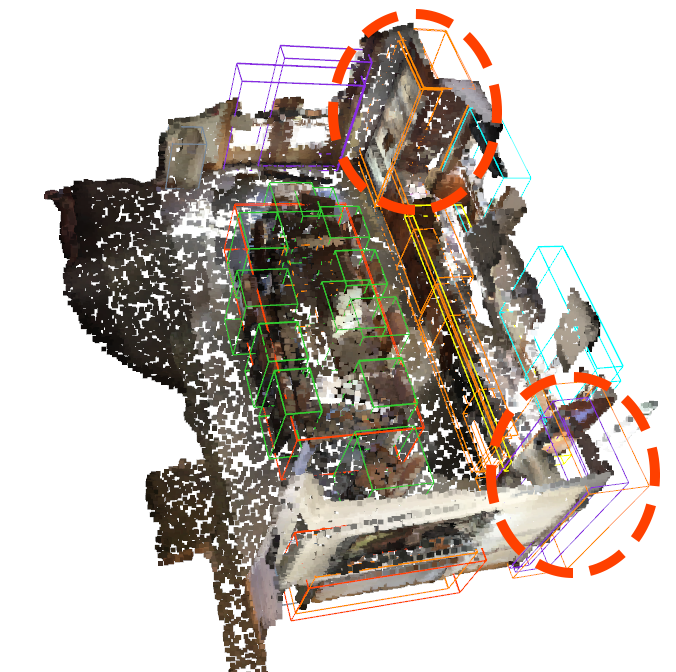}
         &
        \includegraphics[width=0.2\linewidth]{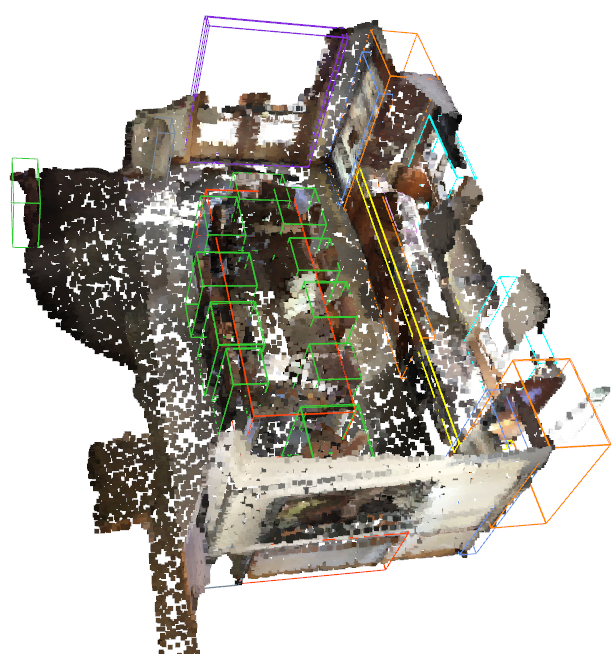}\\
         \includegraphics[width=0.2\linewidth]{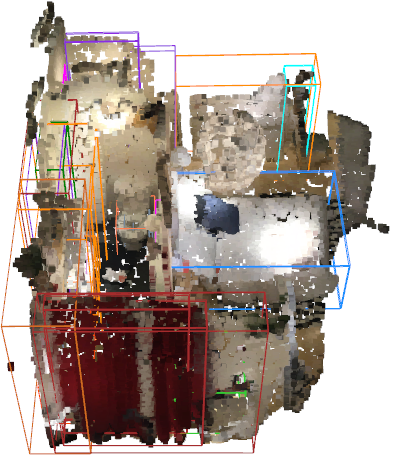}
        &
        \includegraphics[width=0.2\linewidth]{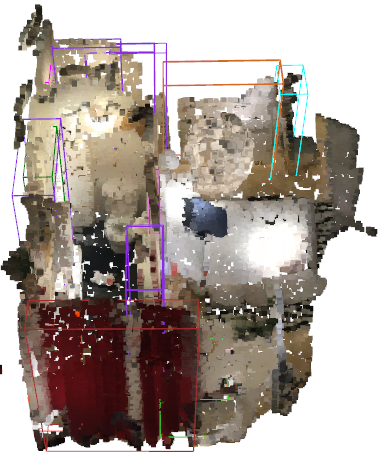}
        &
         \includegraphics[width=0.2\linewidth]{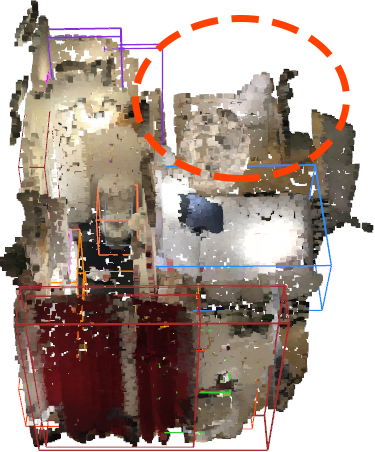}
         &
        \includegraphics[width=0.2\linewidth]{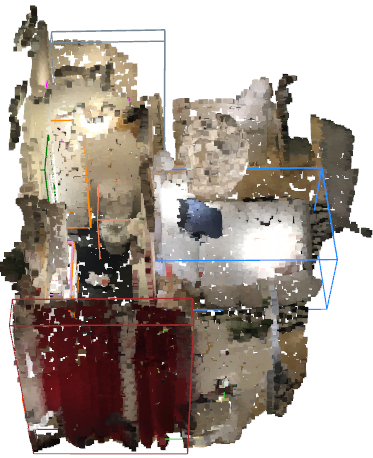}\\
    \end{tabular}
    \end{minipage}
    \begin{minipage}{.2\textwidth}
     \begin{tabular}{cc}  
      \includegraphics[width=0.6\linewidth]{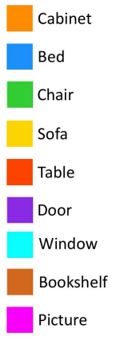} &
      \includegraphics[width=0.65\linewidth]{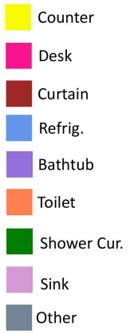}
    \end{tabular}
    \end{minipage}
    \caption{Qualitative results of 3D object detection results on  ScanNetV2 dataset are shown by comparing our models with the baseline model . In the first scene, CMR3D perfectly detects the cabinet highlighted in orange circle. In the second scene, there is a false detection by other models, CMR3D removes the false detection compared to baseline. Note that point clouds are colored only for  illustration.}
    \label{fig:qs}
\end{figure*}
\begin{figure*}
\centering
\begin{minipage}{.5\textwidth}
  \centering
  \includegraphics[width=.95\linewidth]{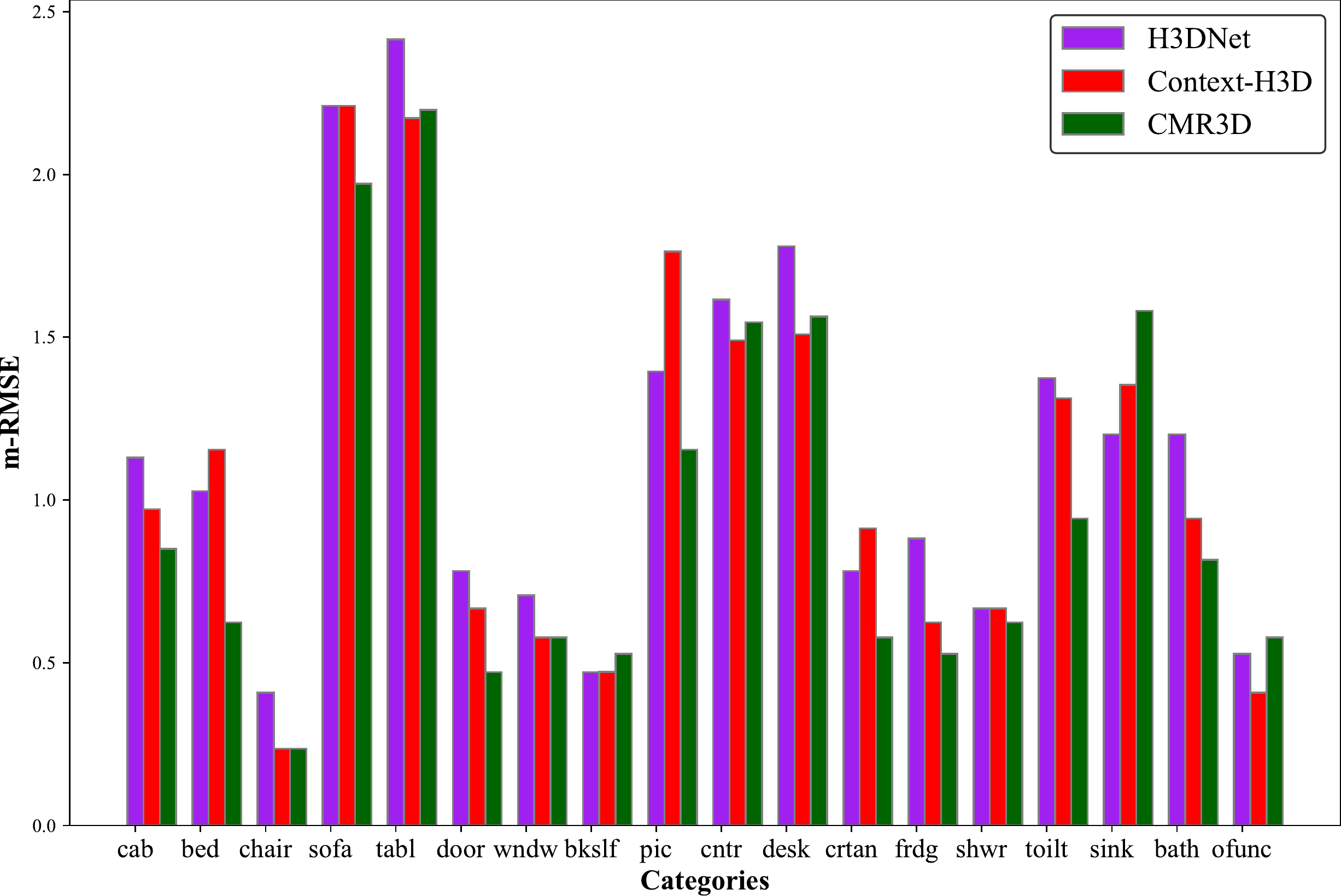}
\end{minipage}%
\begin{minipage}{.5\textwidth}
  \centering
  \includegraphics[width=.95\linewidth]{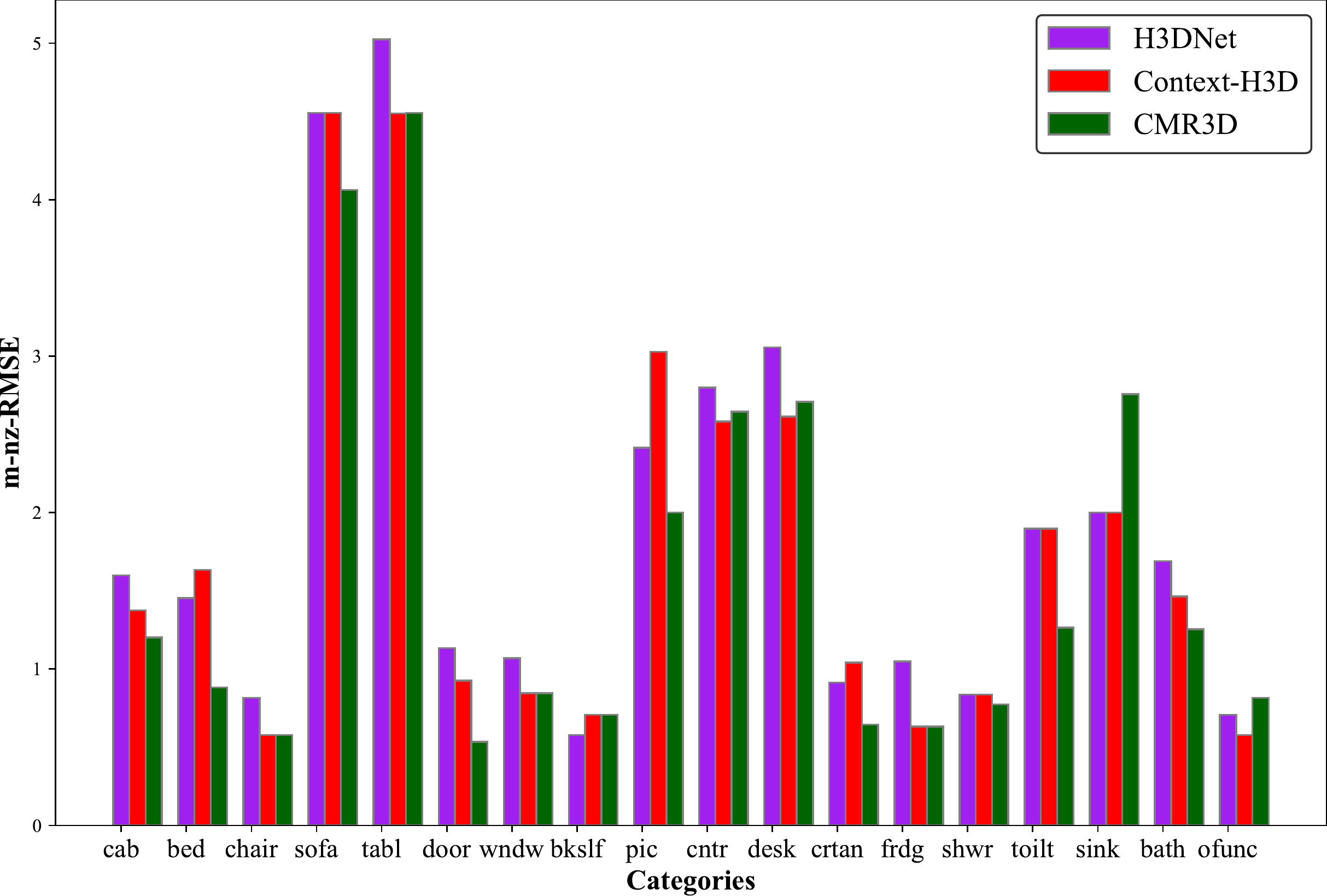}
\end{minipage}

\begin{minipage}{.5\textwidth}
  \centering
  \includegraphics[width=.95\linewidth]{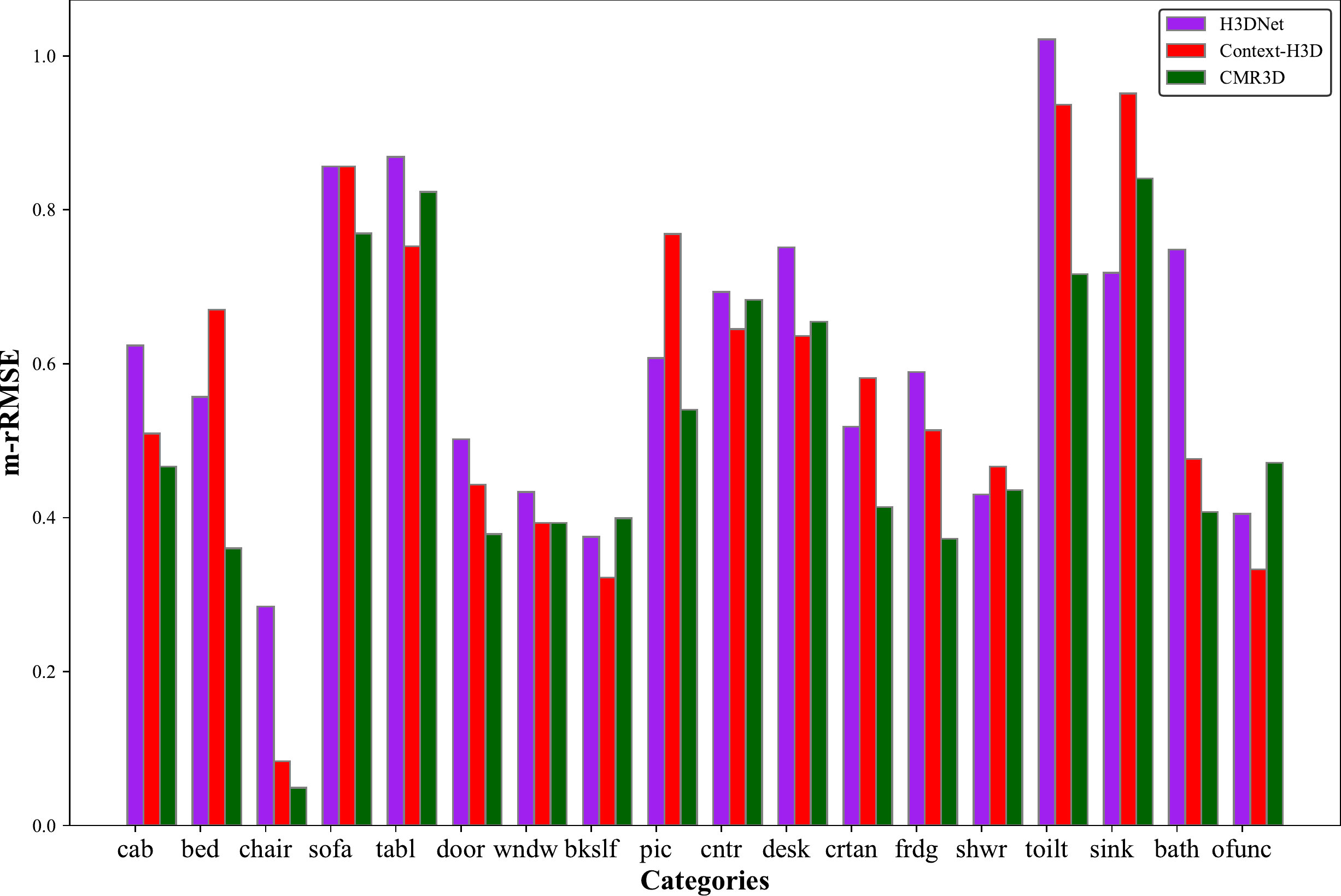}
\end{minipage}%
\begin{minipage}{.5\textwidth}
  \centering
  \includegraphics[width=.95\linewidth]{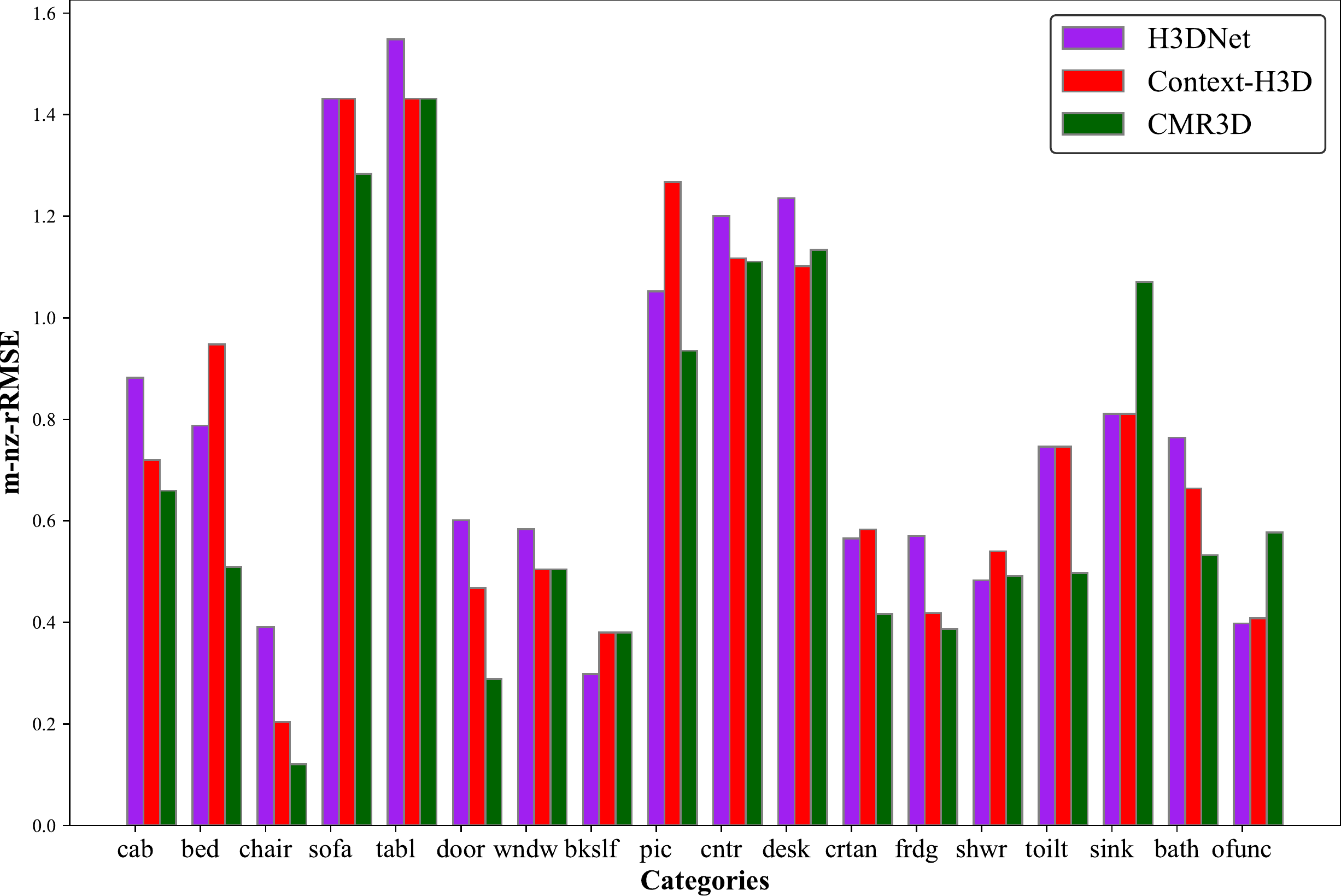}
\end{minipage}
\caption{Category specific object counting comparison on the ScanNet dataset. We show the category specific m-RMSE, m-nz-RMSE, m-rRMSE and m-nz-rRMSE values. \textit{Lower the error value the better.}}\label{rel_rmse_values}
\end{figure*}
\subsection{Ablation Study}
We conduct ablation experiments to understand the effectiveness of the contextual modules.
In Table \ref{tab:modules}, we can observe that introducing each contextual module shows a slight increment in mAP $0.25$ and almost $1\%$ increase in mAP $0.50$. The GCM module alone provides $1\%$ of increment in mAP $0.50$ which clearly provides the evidence that contextual information helps the model in better understanding the scenes. This shows that the contextual modules are helping the model to learn better feature representations. Table \ref{tab:ablation} shows that our CMR3D performs better with single backbone, as it provides a significant increase of about $3\%$ in mAP $0.25$ and $5\%$ in mAP $0.50$. The average mAP of five runs follows the same trend indicating that contextualized multi-stage refinement is helping the model better understand the scenes. We provide few qualitative results which shows that our CMR3D is better in localizing BB and predicting their classes.

\subsection{Qualitative Results} \label{quality}

The qualitative results also helps in better understanding the performance of the model. We have also provided these in order to prove that the contextual information is helping the model to detect bounding boxes accurately. Comparison of the baseline (H3DNet) method with the context enhanced (Context-H3D) and multi-stage refinement (CMR3D) methods is shown in Fig. \ref{fig:qs}.
In the first scene, CMR3D detects cabinets on right side which other methods fail to identify. In second scene, there is a false detection in baseline method however CMR3D does not detect it.  It is observed that integrating contextual modules to baseline method is showing an improvement in predicting the objects, however, with multi-stage refinement, the model is able to detect and classify object category accurately along with refining the 3D bounding box position.

\textbf{RMSE Plots:} Fig. \ref{rel_rmse_values} shows the plot of the $m-RMSE$ values along with the variants such as $m-rRMSE$ $m-nz-RMSE$ and $m-nz-rRMSE$. We found that our CMR3D shows less error compared to other models. This indicates that contextual information is enhancing the performance of our model. However, we notice that the sink category has more error obtained by our model compared with the baseline method in $m-RMSE$ and $m-nz-RMSE$ plots.

\section{Conclusion}
We propose a contextualized multi-stage refinement for 3D object detection, where we capture the contextual information at different granularity levels. We utilize self-attention and multi-levels of refinement to capture the relationship between the objects and their surroundings and refine the location of object. We also show the benefits of our proposed method for object counting. 
For future work, variations of the self-attention modules can be employed to retrieve contextual information.

{\small
\bibliographystyle{ieee_fullname}
\bibliography{egbib}
}

\end{document}